\documentclass[preprint,12pt]{elsarticle}

\usepackage{amssymb}
\usepackage{amsmath}
\usepackage{siunitx}
\usepackage[hyphens]{url}      
\usepackage[hidelinks]{hyperref}
\usepackage{doi}               

\usepackage{enumitem} 
\usepackage{graphicx}
\usepackage{subcaption}

\usepackage{makecell}
\usepackage{booktabs}
\usepackage{multirow}
\usepackage{placeins}
\biboptions{sort&compress}

\usepackage[acronym,nonumberlist]{glossaries}
\makenoidxglossaries
\newacronym{soc}{SOC}{State-of-Charge}
\newacronym{spm}{SPM}{Single Particle Model}
\newacronym{spme}{SPMe}{SPM with electrolyte dynamics}
\newacronym{dfn}{DFN}{Doyle-Fuller-Newman}
\newacronym{pde}{PDE}{Partial Differential Equation}
\newacronym{pinn}{PINN}{Physics-Informed Neural Network}
\newacronym{deeponet}{DeepONet}{Deep Operator Network}
\newacronym{pideeponet}{PI-DeepONet}{Physics-Informed DeepONet}
\newacronym{fno}{FNO}{Fourier Neural Operator}
\newacronym{pefno}{PE-FNO}{parameter-embedded Fourier Neural Operator}
\newacronym{mlp}{MLP}{Mixed Layer Perceptron}

\newglossaryentry{lfp}{name=LFP, description={Lithium Iron Phosphate}}
\newglossaryentry{cpu}{name=CPU, description={Central Processing Unit}}
\newglossaryentry{gpu}{name=GPU, description={Graphics Processing Unit}}

\journal{Energy and AI}

\begin{document}

\begin{frontmatter}

\title{Fast and Generalizable parameter-embedded Neural Operators for Lithium-Ion Battery Simulation}

\author[carl,isea,imperialmech,dyson]{Amir Ali Panahi,\corref{cor1}\fnref{eq}}
\author[carl,isea]{Daniel Luder,\fnref{eq}}
\author[dyson,faraday]{Billy Wu}
\author[imperialmech,faraday,vfh]{Gregory Offer}
\author[carl,isea]{Dirk Uwe Sauer}
\author[carl,isea]{Weihan Li, \corref{cor2}}

\cortext[cor1]{Corresponding author.  E-mail: a.panahi25@imperial.ac.uk}
\cortext[cor2]{Corresponding author.  E-mail: weihan.li@isea.rwth-aachen.de}
\fntext[eq]{Equal contributions.}

\affiliation[carl]{%
  organization={Center for Ageing, Reliability and Lifetime Prediction of Electrochemical and Power Electronic Systems (CARL), RWTH Aachen University},
  addressline={Campus-Boulevard 89},
  city={Aachen},
  postcode={52074},
  country={Germany}
}

\affiliation[isea]{%
  organization={Institute for Power Electronics and Electrical Drives (ISEA), RWTH Aachen University},
  addressline={Campus-Boulevard 89},
  city={Aachen},
  postcode={52074},
  country={Germany}
}

\affiliation[imperialmech]{%
  organization={Department of Mechanical Engineering, Imperial College London},
  city={London},
  postcode={SW7 2AZ},
  country={United Kingdom}
}

\affiliation[dyson]{
  organization={Dyson School of Design Engineering, Imperial College London},
  addressline={South Kensington Campus},
  city={London},
  postcode={SW7 2AZ},
  country={United Kingdom}
}

\affiliation[faraday]{%
  organization={The Faraday Institution},   
  addressline={Harwell Science and Innovation Campus},
  city={Didcot},
  postcode={OX11 0RA},
  country={United Kingdom}
}

\affiliation[vfh]{%
  organization={Vehicle Futures Hub, Imperial College London},
  city={London},
  postcode={SW7 2AZ},
  country={United Kingdom}
}



\begin{abstract}
Reliable digital twins of lithium‑ion batteries must achieve high physical fidelity with sub‑millisecond speed. In this work, we benchmark three operator‑learning surrogates for the \acrfull{spm}: \acrfullpl{deeponet}, \acrfullpl{fno} and a newly proposed \acrfull{pefno}, which conditions each spectral layer on particle radius and solid‑phase diffusivity. Models are trained on simulated trajectories spanning four current families (constant, triangular, pulse‑train, and Gaussian‑random‑field) and a full range of \acrfull{soc} (\SIrange{0}{100}{\%}). \acrshort{deeponet} accurately replicates constant‑current behaviour but struggles with more dynamic loads. The basic \acrshort{fno} maintains mesh invariance and keeps concentration errors below \SI{1}{\%}, with voltage mean‑absolute errors under \SI{1.7}{\milli\volt} across all load types. Introducing parameter embedding marginally increases error, but enables generalisation to varying radii and diffusivities. \acrshort{pefno} executes approximately 200 times faster than a 16‑thread \acrshort{spm} solver. Consequently, \acrshort{pefno}'s capabilities in inverse tasks are explored in a parameter estimation task with Bayesian optimisation, recovering anode and cathode diffusivities with \SI{1.14}{\%} and \SI{8.4}{\%} mean absolute percentage error, respectively, and \si{0.5918} percentage points higher error in comparison with classical methods.
These results pave the way for neural operators to meet the accuracy, speed and parametric flexibility demands of real‑time battery management, design‑of‑experiments and large‑scale inference. \acrshort{pefno} outperforms conventional neural surrogates, offering a practical path towards high‑speed and high-fidelity electrochemical digital twins.
\end{abstract}








\begin{keyword}
Physics-informed machine learning \sep Operator Learning \sep Deep Operator Network \sep Fourier Neural Operator \sep Lithium-Ion Batteries


\end{keyword}

\end{frontmatter}

\section{Introduction}
\label{intro}

Physics-based models of lithium-ion batteries have become indispensable for electrified transportation, grid energy storage and second-life applications. These models support diverse tasks such as fast-charging optimisation~\cite{zou_model_2018}, design of experiments~\cite{roman-ramirez_design_2022} and health diagnostics~\cite{roman_machine_2021, xiong_towards_2018}.

\medskip
\noindent
By describing mass-, charge-, and energy-conservation within the porous electrodes and electrolyte, battery models reduce to systems of coupled diffusion–migration and charge-transfer \acrfullpl{pde}. The \acrfull{dfn} model remains the gold standard for such physics-based modeling. However, solving its coupled \acrshortpl{pde} still takes \SIrange{3}{115}{\milli\second} per cycle, even with the fastest open-source solvers under isothermal constant-current conditions, making it challenging for tasks requiring broad condition coverage, degradation analysis, or rapid parameter sweeps.~\cite{doyle_modeling_1993,berliner_methodspetlion_2021}

\medskip
\noindent
To enable battery digital twins that can operate alongside physical cells and inform real-time decisions, further acceleration beyond the \acrshort{dfn} is necessary. Conventional strategies address this by developing surrogate models that retain the governing physics while simplifying geometric or dimensional complexity. The \acrfull{spm}, and its electrolyte-enhanced variant \acrfull{spme}, exemplify this strategy by homogenising electrode microstructure and simplifying electrolyte dynamics~\cite{marquis_asymptotic_2019}. Nevertheless, reliance on spatio-temporal grids often lead to computational bottlenecks, particularly when numerous forward simulations are needed—for example, in large-scale parameter estimation~\cite{hong_improved_2023} or embedded model-predictive control~\cite{hwang_model_2022}. In parallel, data-driven methods have become omnipresent in battery research, delivering advances in \acrfull{soc} and state-of-health estimation, fast-charging strategies, closed-loop experimentation, and design optimisation~\cite{attia_closed-loop_2020,li_digital_2020,yuan_comparative_2025,li_online_2021,gayon_lombardo_machine_2021}. Yet, despite their success, purely data-driven models often struggle to accurately extrapolate beyond their training domain and provide limited physical interpretability, making data-driven models indispensable for safety-critical applications~\cite{borah_synergizing_2024}.

\medskip
\noindent
A promising direction is to embed physical knowledge directly into machine learning architectures. This approach, often referred to as physics-informed, physics-inspired or scientific machine learning, seeks to combine neural networks with electrochemical theory to achieve the simulation speed of data-driven surrogates while retaining physical interpretability. 


\medskip
\noindent
Within scientific machine learning, several strategies now incorporate physical constraints into neural networks. Most prominently, \acrfullpl{pinn} embed the governing \acrshort{pde} residuals into the loss function by leveraging automatic differentiation in deep learning frameworks. This ensures that the learned solution adheres to the same physical laws as the original system~\cite{chen_universal_1995,raissi_physics-informed_2019}. In battery modelling, \acrshortpl{pinn} have been successfully applied to parameter estimation~\cite{wang_physics-informed_2024}, state estimation~\cite{li_physics-informed_2021,wang_physics-informed_2022}, long-horizon health prognosis~\cite{hofmann_physics-informed_2023,sun_adaptive_2023} and thermal modelling~\cite{kim_modeling_2023,pang_physics-informed_2023}. Yet, their reliance on specific boundary conditions, such as the applied current in physics-based models, often necessitates custom retraining or costly fine-tuning when drive cycles change. When the governing physics are only partially known, another strategy of scientific machine learning, coined Universal Differential Equations, introduces learnable correction terms to known physical models, allowing neural components to fill gaps in the physics. These hybrid differential equations can be trained end-to-end using experimental data to augment incomplete mechanistic knowledge~\cite{rackauckas_universal_2021}. This paradigm has been used in~\cite{kuzhiyil_neural_2024,nascimento_hybrid_2021} to increase modelling accuracy of reduced-order models and in~\cite{kuzhiyil_lithium-ion_2025} to model ageing modes such as solid electrolyte interphase (SEI)-layer growth and pore clogging.

\medskip
\noindent
In a similar vein, computationally challenging parts of the battery \acrshortpl{pde}, such as root-finding in the algebraic equation of the \acrshort{dfn} model, have been replaced by neural networks, allowing for higher speed~\cite{huang_minn_2024}. Consequently, the idea has been raised to replace simulation altogether, by directly learning the solution operator of the \acrshort{pde} itself. This idea, known as operator learning, uses classical solvers to obtain simulation data of the \acrshort{pde} of interest, based on many different boundaries, initial conditions and parameters and learns the operator described by this mapping in function space~\cite{kovachki_neural_2023}. 

\medskip
\noindent
Among various operator-learning architectures developed to date, two stand out as foundational: \acrfull{deeponet} and \acrfull{fno}. \acrshort{deeponet} represents the input field as a vector of sensor values, interprets these as coefficients in a learnable function space, and computes the output through an inner product between this coefficient vector and a set of basis functions generated by a trunk-network at the query point~\cite{lu_learning_2021}. 

\medskip
\noindent
\acrshortpl{deeponet} can be extended to incorporate physical constraints by embedding residual terms from the governing equations into the loss function, yielding \acrfullpl{pideeponet}. This approach preserves the flexibility of operator learning while imposing physical laws. For instance, \citet{zheng_inferring_2023} demonstrated a composite surrogate for a polynomial-electrolyte \acrshort{spm} using three \acrshortpl{deeponet}: two model the mapping from current to concentration for the positive and negative electrodes, while a third maps the resulting concentration fields to terminal voltage. Although replacing the algebraic voltage relation with a neural mapping is arguably unnecessary. To further improve physical interpretability, solid-phase diffusivities are introduced as explicit inputs to each branch network and treated as constant parameters. Embedding a residual physics loss in the concentration networks improves accuracy by enforcing consistency with the \acrshortpl{pde}. Because the full surrogate is differentiable, it supports gradient-based recovery of diffusivities directly from voltage data. However, the model remained constrained in scope, having been trained solely on linearly ramped single charge-discharge cycles at a fixed \acrshort{soc}, limiting its generalisation to broader operating conditions. The same authors later reformulated their approach as a state-space propagator~\cite{zheng_state-space_2023}. Here, a single \acrshort{pideeponet} advances the battery states over time, enabling simulations from arbitrary initial \acrshortpl{soc} and thus relaxing prior constraints on operating conditions. This gain in generality, however, comes at the expense of input diversity; the network is trained and evaluated solely under constant-current profiles and for a fixed parameter set. To address this limitation,  \citet{brendel_parametrized_2025} extended the method to cover a wide range of applied currents by training \acrshortpl{pideeponet} directly on solutions of the \acrshort{spm} equations, bypassing the need for synthetic or experimental data. Their model incorporates scalar diffusivities as inputs via the trunk network, while spatial–temporal coordinates are lifted into a higher-dimensional feature space using multi-scale spatio-temporal Fourier-feature embeddings. This enables the network to handle a broad, two-decade range of diffusivity values. The resulting surrogate allows for efficient computation of Fisher information matrices, accelerating Design-of-Experiments workflows by facilitating fast gradient evaluations. A key finding across these efforts is that improved generalisation, with respect to both state and parameter spaces, requires significantly longer training times, typically in the range of \SIrange{6}{23}{\hour}~\cite{zheng_inferring_2023, brendel_parametrized_2025}.

\medskip
\noindent
In contrast to \acrshortpl{deeponet}, the \acrshort{fno} framework assumes that the underlying operator is realised through integrating with a translation-invariant convolutional kernel, reducing the problem to multiplication in Fourier space. This approach avoids explicit learning of the operator across spatial locations, as the kernel values are directly learned in frequency space. By transforming the input into Fourier space, applying the learned kernel, and then performing an inverse transform, \acrshortpl{fno} achieve a spectral convolution across the geometry~\cite{li_fourier_2021}. Notably, \acrshortpl{fno} enable mesh-free extrapolation by learning the operator at coarse resolution and then inferring fine-mesh solutions, which is a property critical for computational efficiency. As a result, \acrshortpl{fno} have been classified as “true” neural operators \cite{kovachki_neural_2023}, while  \acrshortpl{deeponet} are considered neural regression surrogates due to their pointwise nature.

\smallskip
\noindent
The basic \acrshort{fno} has been further extended by a parameter-embedding logic first introduced by \citet{takamoto_learning_2023}, which allows the model to handle particle-dependent diffusivity and radius parameters. These scalar inputs are passed through a \acrfull{mlp}, producing channel-wise modulation factors that scale the lifted input. Each Fourier layer is conditioned on these parameters, while the core architecture and learned convolution kernel remain unchanged. This design enables the model to generalize across different particle characteristics without retraining.

\medskip
\noindent
To the authors’ best knowledge, no prior work has applied the \acrshort{fno} to learn the solution operator of battery \acrshortpl{pde}. The only related study \cite{kwak_robust_2023} repurposed the architecture as a lithium-ion battery state estimator; an application that is fundamentally distinct due to the ill-posed nature of state estimation and the estimator’s mathematical formulation. The main contributions of this work are as follows:
\begin{enumerate}[label=(\roman*)]
  \item The introduction of the first neural-operator surrogate that captures the full spatio-temporal solution of an electrochemical lithium-ion battery model, independent of any fixed mesh or time discretization.
  \item A compact, physically motivated parameter-embedding scheme that maintains surrogate accuracy across a broad range of \acrshortpl{soc}, dynamic current profiles, particle radii, and solid-phase diffusion coefficients.
  \item Comprehensive timing evaluations demonstrating one to two order of magnitude speed-up over state-of-the-art multithreaded \acrshort{spm} solvers, while maintaining high fidelity.
  \item Deployment of the surrogate within a Bayesian optimization-based parameter estimation task, providing the first analysis of how forward-model errors propagate to inverse-problem accuracy.
\end{enumerate}

\section{Methods}
\label{sec:methods}

\subsection{Electrochemical Model}

\noindent
In this work, the \acrshort{spm} described in \cite{marquis_asymptotic_2019} is used, where the assumptions are made that the electrical conductivity of the electrolyte is sufficiently high, and the timescale for lithium ion migration within the electrolyte is negligible compared to the typical timescale of a charge or discharge. Consequently, the electrolyte concentration is treated as constant. The battery dynamics are modelled section-wise in the negative and positive domains. As such, the index $k \in \{n, p\}$ refers to the domain of interest. Lithium transport within each particle is governed by the diffusion equation:

\begin{subequations}\label{eq:spm}
    \begin{align}
            \frac{\partial c_k}{\partial t}
        &+\frac{1}{r^{2}}\frac{\partial}{\partial r}\!\bigl(r^{2}N_k\bigr)=0,
        \label{eq:spm_a}\\[2pt]
        N_k&=-D_k(c_k)\,\frac{\partial c_k}{\partial r},
        \qquad 0\le r\le R_k,\;k\in\{n,p\},
        \label{eq:spm_b}
    \intertext{Here, \(c_k\) is the solid-phase lithium concentration in electrode \(k\). For simplicity, we drop the additional subscript \(s\) and use \(c_k\) throughout. The solid-phase diffusivity \(D_k(c_k)\) is taken as constant in this work, \(D_k(c_k)\equiv D_k\).
    The dimensionless radii and concentrations are defined as \(r'_k=r/R_k\) and \(c'_k=r/c_{k,\text{max}}\), respectively, where $c_{k,\text{max}}$ is the maximum lithium concentration in the respective electrode. The cell voltage then equates to:}
            \label{eq:spm_c} 
            V(t) &= \underbrace{U_p(c'_{p})\big|_{{r'_p}=1} - U_n(c_{n}')\big|_{{r'_n}=1}}_{\text{Open circuit voltage}} \\
       & \quad - \underbrace{\frac{2RT}{F} \left( \sinh^{-1}\left(\frac{I(t)}{a_p j_{0,p} L_p}\right)
       + \sinh^{-1}\left(\frac{I(t)}{a_n j_{0,n} L_n}\right)\right)}_{\text{Reaction overpotentials}}, \nonumber \\
    \intertext{where \(U_k(\,\cdot\,)\) denotes the equilibrium potential of electrode \(k\) as a
    function of surface stoichiometry, \(R\) is the universal gas constant, \(T\) is the cell temperature
    and \(F\) is Faraday’s constant.  The key geometric/electrode parameters are:
            \begin{itemize}
          \item \(L_k\): thickness of electrode \(k\),
          \item \(a_k\): specific interfacial area of electrode \(k\).
        \end{itemize}
        The exchange-current density is modelled as}
        j_{0,k}&=\sqrt{c'_k\bigl(1-c'_k\bigr)}\Bigg|_{{r'_k}=1},\qquad k\in\{n,p\}.
    \intertext{The model's boundary and initial conditions are given by:}
        N_k&=0,\qquad\qquad\qquad\text{on }r=0,\label{eq:spm_d}\\
        N_k&=
        \begin{cases}
          \dfrac{IR_k}{3\,\varepsilon_k F L_k A}, & \text{on }r=R_k,\;k=n,\\[6pt]
          -\dfrac{IR_k}{3\,\varepsilon_k F L_k A}, & \text{on }r=R_k,\;k=p,
        \end{cases}\label{eq:spm_e}\\
        c_k&(r,0)=c_{k0}(r),\qquad k\in\{n,p\},\label{eq:spm_f}
    \end{align}
\end{subequations}
with \(I>0\) representing current flowing \emph{into} the cell (charging).  Here \(c_{k0}(r)\) is the initial uniform solid concentration in electrode \(k\) and the geometric/electrode parameters are:
\begin{itemize}
  \item \(A\): cell cross-sectional area,
  \item \(\varepsilon_k\): solid-phase volume fraction in electrode \(k\).
\end{itemize}

\begin{figure}[htb]
    \centering
    \includegraphics[width=0.5\columnwidth]{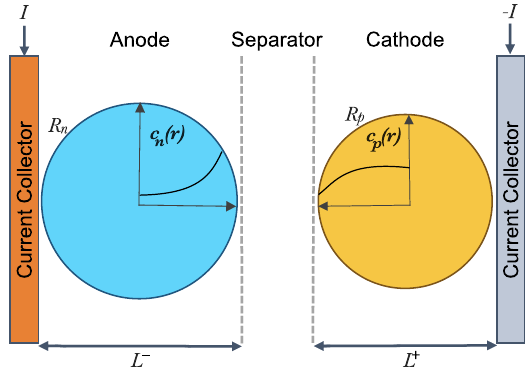}
    \caption{Single Particle Model during a charging process}\label{fig:spm}
\end{figure}

\noindent
A schematic of the the \acrshort{spm} is visualised in Figure \ref{fig:spm}. In this work, we employ a LiFePO$_4$--graphite pouch cell parametrised by \citet{prada_simplified_2013} as the reference chemistry. For the negative electrode, where the OCP curve is not reported, values are adopted from the LG M50 NMC811 cell characterization of \citet{chen_development_2020}. Parameter sets for both chemistries are given in the \nameref{app}. Although LiFePO$_4$ is known to lithiate via a two-phase mechanism rather than the solid-solution behaviour assumed by the \acrshort{spm}, we retain the conventional formulation here because the objective of this paper is to benchmark the estimation methodology; this simplification does not influence the methodological insights that follow.

\subsection{Operator Learning}
\label{sec:op_learning}
For each electrode \(k\in\{n,p\}\), let
\(X_k=[0,R_k]\times[0,T]\subset\mathbb{R}^2\) denote the
space–time domain.
Given an applied current profile \(I:[0,T]\!\to\!\mathbb{R}\) and an
initial concentration \(c_{0,k}:[0,R_k]\!\to\!\mathbb{R}\), we define
\(
   a=(I,c_{0,k})\in
   \mathcal{A}=\mathcal{I}\times\mathcal{C}_0,
\)
where \(\mathcal{I}\) and \(\mathcal{C}_0\) are the corresponding function
spaces. For each \(k \in \{n,p\}\), an operator is learned:
\[
   G_k:\mathcal{A}\longrightarrow
   \mathcal{C}_k=C(X_k),\qquad
   c_k(\cdot,\cdot)=G_k(I,c_{0,k}),
\]
where \(c_k(r,t)\) solves
Equations \eqref{eq:spm_a}–\eqref{eq:spm_f} on \(X_k\).
We train an electrode-specific surrogate
\(G_{\theta_k}\approx G_k\) by minimising their normalized error on the $L_2$-norm over the entire electrode domain and simulation horizon:
\begin{equation}
   \min_{\theta_k\in\Theta}\,
   \frac1N\sum_{j=1}^{N}
   \frac{\|G_{\theta_k}(I_j,c_{0,k,j})-G_k(I_j,c_{0,k,j})\|_{L_2(X_k)}}
        {\|G_k(I_j,c_{0,k,j})\|_{L_2(X_k)}}.
   \label{eq:l2_rel_k}
\end{equation}
where \(N\) denotes the number of training samples.

\medskip
\noindent
To ensure that we operate within ranges where the \acrshort{spm} itself accurately captures the underlying physics, the current profiles are clipped to \(-1.5C \leq I(t) \leq 1.5C\).

\begin{figure*}[!htb]
  \begin{subfigure}[t]{0.3694\textwidth}
    \includegraphics[width=\linewidth]{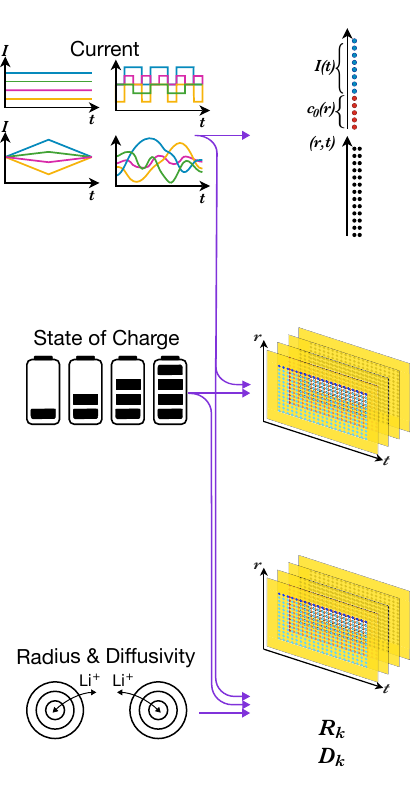}
    \subcaption{}\label{fig:net_a}
  \end{subfigure}%
  \begin{subfigure}[t]{0.371\textwidth}
    \includegraphics[width=\linewidth]{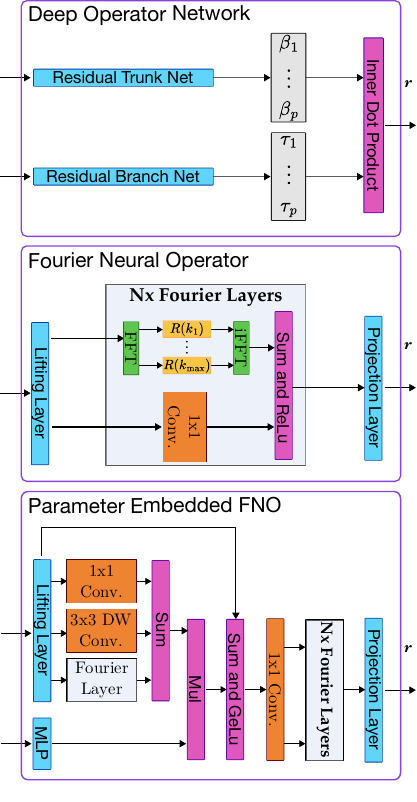}
    \subcaption{}\label{fig:net_b}
  \end{subfigure}%
  \begin{subfigure}[t]{0.2585\textwidth}
    \includegraphics[width=\linewidth]{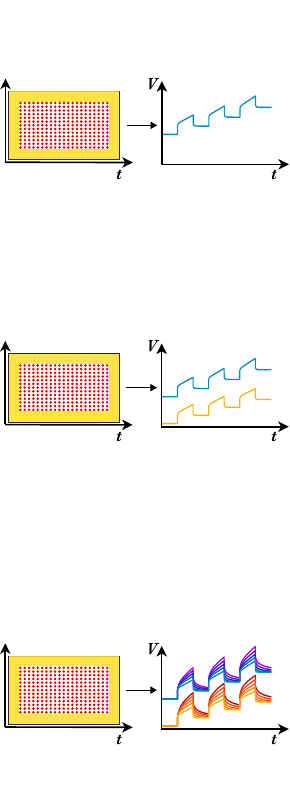}
    \subcaption{}\label{fig:net_c}
  \end{subfigure}

  \caption{Overview of the architectures.  
(a) \emph{Input design.}  The applied current \(I(t)\) and the
initial solid concentration \(c_{0,k}(r)\) are broadcast to a common
\((r,t)\) grid; yellow margins indicate zero–padding cells.  
For the \acrshort{pefno} the scalar diffusivity \(D_k\) and particle radius $R_k$ is fed to an \acrshort{mlp} at this
stage.  
(b) \emph{Operator networks.}  
\textbf{\acrshort{deeponet}:} identical residual branch/trunk nets produce basis
coefficients and functions that are combined via a coefficient–basis
sum.  
\textbf{\acrshort{fno}:} a stack of \(N\) Fourier layers with a \(1{\times}1\)
channel-wise residual path processes the lifted tensor.  
\textbf{\acrshort{pefno}:} a parameter-embedding block (\(1{\times}1\) conv, depth-wise
\(3{\times}3\) conv, single Fourier layer) is multiplied to a feature lifted $D_k$ before the standard Fourier stack.  
(c) \emph{Projection head.}  A final linear projection returns the
predicted concentration field \(c_k(r,t)\), which is subsequently
converted to terminal voltage \(V(t)\).}
  \label{fig:input_architecture_output}
\end{figure*}

To cover and extend earlier current spaces
\cite{zheng_inferring_2023,brendel_parametrized_2025}, we sample four families:
\begin{itemize}
  \item \textbf{CC}: constant current with random amplitude,
  \item \textbf{TRI}: triangular ramp \((0\!\to\!\pm I_{\max}\!\to\!0)\),
  \item \textbf{PLS}: rectangular pulse trains with random pulse count, duty cycle and sign,
  \item \textbf{GRF}: smoothly varying profile from a periodic Gaussian random field.
\end{itemize}

\noindent
Exact sampling rules are summarised in the \nameref{app}.
Training pairs \(\{(I_j,c_{0,k,j}),\,c_{k,j}\}_{j=1}^N\) are generated with \texttt{PyBaMM}~\cite{sulzer_python_2021}.
Because the negative and positive particles interact only through the shared input current, independent learning of
\(G_n\) and \(G_p\) is both natural and computationally convenient.

\paragraph{\acrshort{deeponet}}
In the \acrshort{deeponet} framework, the operator learning task is realised by combining two components: a \emph{branch net} that learns feature-dependent basis coefficients $\{\beta_i(a)\}_{k=1}^p$ for the current–concentration pairs, and a \emph{trunk net} that encodes geometry dependent basis functions $\{\tau_i(r,t)\}$ for spatio-temporal sampling points.
Their inner product $\langle\beta(a),\tau(r,t)\rangle$
is itself a function of $a$ evaluated on $(r,t)$ and predicts the concentration:
\begin{equation}
   c_k(r,t)\approx G_{\theta_k} (a)(r,t)=\sum_{i=1}^{p}\beta_{k,i}(a)\tau_{k,i}(r,t),\; k \in \{n,p\}.
\end{equation}

\paragraph{\acrshort{fno}}
The Fourier Neural Operator treats the discretised input
\[
   a(r,t)=\bigl[I(t),\;c_{0,k}(r),\;r/R_k,\;t/T\bigr] \in\mathbb{R}^{4}
\]
as a four-channel image over the \((r,t)\) grid.\,%
A linear lifting layer \(P\) maps each grid point to a \(d\)-dimensional latent vector \(v^{0}=P\,a\).
An \emph{FNO block} then performs a spectral convolution combined with a point-wise linear update:
\begin{equation}
   v^{\ell+1}(x)=
      \sigma\!\Bigl(
        W\,v^{\ell}(x)\;+\;
        \mathcal{F}^{-1}\!\bigl(
          R_\phi \,\odot\,\mathcal{F}v^{\ell}
        \bigr)\!(x)
      \Bigr),
      \; \ell=0,\dots,L-1,
\end{equation}
where \(\mathcal{F}\) and \(\mathcal{F}^{-1}\) are the forward and inverse Fourier transforms, \(R_\phi\in\mathbb{C}^{k_{\max}\times d\times d}\) is the kernel representation in frequency space storing learnable complex weights for the lowest \(k_{\max}\) modes, and \(\odot\) denotes element-wise multiplication. The pointwise weight matrix \(W \in \mathbb{R}^{C_{\text{out}}\times C_{\text{in}}}\) act as a \(1\times1\) kernel, mixing channels identically at every grid point and carrying local information through the network. In combination with the activation function \(\sigma\), this design captures both low-frequency global correlations (via the spectral convolution) and high-frequency content.

By stacking \(L\) such \acrshort{fno} blocks, we obtain a latent field \(v^{L}\). A final projection \(Q\) maps this latent field back to a single channel and gives the prediction for $c_k(r,t)$.

\paragraph{\acrshort{pefno}}
To cover the family of
\acrshort{spm} problems that arise under varying solid particle diffusivity and radii parameters, we adopt
the parameter-embedding strategy of~\citet{takamoto_learning_2023}.

In this framework, three parallel branches are applied to the lifted input of the \acrshort{fno}:

\begin{enumerate}[label=(\alph*)]
  \item \label{itm:mix} a \(1\times1\) convolution that mixes channels;
  \item \label{itm:dw}  a depth-wise (DW) \(3\times3\) convolution that injects local-neighbourhood information;
  \item \label{itm:fourier} a Fourier layer that supplies the first global interaction.
\end{enumerate}
  
The three outputs are summed and then modulated by the material parameters by multiplication with their latent-space representations through a two-layer \acrshort{mlp}.

\smallskip
\noindent
The \(3{\times}3\) depth-wise convolution allows the modulation to influence each node through its immediate neighbours. Mathematically, the layer processes the neighbourhood in a manner analogous to a finite-difference method of a second-order derivative, making it a natural fit for modelling diffusion processes.

Distinct from that, the \(1{\times}1\) convolution and the Fourier layer allow the parameters to affect local and global effects between the channels. This design ensures that geometric parameters, such as particle radius, can influence both the boundary channel at $r=R_k$ and the
interior fields in one stroke, without further architectural adjustments. Together, these mechanisms provide the network with the flexibility to accurately represent the physical effects of varying \(D_k,\,R_k\) without altering the core \acrshort{fno}.

Note that formally, the input to the \acrshort{pefno} becomes $a = (I,c_{0,k},D_k,R_k) \in \mathcal{I}\times\mathcal{C}_0\times \mathcal{D}\times \mathcal{R}$ where $\mathcal{D},\,\mathcal{R}$ are the respective parameter spaces. Figure~\ref{fig:net_a} sketches the resulting input spaces and their respective flow into the three architectures, which are themselves illustrated in Figure~\ref{fig:net_b}.

\subsection{Practical Implementation}

All model parameters, including the initial \acrshort{soc}, are sampled with a Sobol sequence. For the parameters, we sample (omitting units) \(D_k\in[1^{-18},1^{-14}]\), $R_n \in [4^{-6},1.5^{-5}]$ and $R_n \in [1^{-8},1.5^{-5}]$, then apply a logarithmic transform and normalize to \([-1,1]\) before passing them to the network. \acrshort{soc} values are rounded to the nearest practically relevant percentage. In this work, we train using \SI{1}{\hour} simulations; anecdotal experiments indicated that different simulation lengths yield similar model performance. 
\smallskip
\noindent
Branch and trunk nets of the \acrshort{deeponet} are identical ResNets
\cite{he_deep_2015}. The number of basis functions is set to
\(p=16\) for \emph{CC}/\emph{TRI} inputs and \(p=64\) for the more irregular \emph{PLS}/\emph{GRF} inputs.
For both \acrshort{fno} variants, every input channel must live on the same \((r,t)\) grid, so the current \(I(t)\) is broadcast along the \(r\)-axis and the initial concentration \(c_{0,k}(r)\) along the \(t\)-axis.
To accommodate non-periodic signals, we zero-pad the input in \((r,t)\) and include the coordinate channels \((r/R_k,\,t/T)\) exactly as in \cite{li_fourier_2021}.
Numerical values for \(k_{\max}\), depth, width, and parameter-embedding modulation are given in Table~\ref{tab:hparams}. For the \acrshort{pefno}, different numbers of frequency modes $(k_r,k_t)$ are kept for each dimension. A schematic of the input design can be seen in Figure \ref{fig:net_a}.

\begin{table}[t]
\centering
\setlength{\tabcolsep}{4pt}       
\caption{Key hyper-parameters for the three architectures.
Empty entries (—) indicate that the setting does not apply.}
\label{tab:hparams}
\renewcommand{\arraystretch}{1.05}
\begin{tabular}{@{}lccc@{}}
\toprule
\textbf{Hyper-parameter} & \textbf{\acrshort{deeponet}} &
\textbf{\acrshort{fno}} & \textbf{\acrshort{pefno}} \\ \midrule
Core width               & 500 & 32 & 64 \\
Core depth               & 11  & 6  & 8  \\
Basis functions $p$      & 16 / 64 & — & — \\[2pt]
$k_{\max}$               & —       & 10 & (5,20) \\[2pt]
Last input dim.          & \makecell{95 (branch)\\20 (trunk)}
                         & 4 & 4 \\[2pt]
PE layer width        & — & — & 32 \\[2pt]
PE layer depth        & — & — & 2 \\[2pt]
Padding $(r,t)$          & — & $(+2,\,+5)$ & $(+2,\,+5)$ \\ \bottomrule
\end{tabular}
\end{table}

The size of each training set is chosen to match the number of physical degrees of freedom the corresponding learned operator has to cover. During the training, it became apparent that \acrshort{deeponet} can attend to different initial concentrations in general, but not to a varying range in one training process. Therefore, \acrshort{deeponet} is trained only for a single diffusivity pair and a fixed initial \acrshort{soc}, so \(2{,}200\) current profiles are sufficient. In contrast, \acrshort{fno} must generalise over a distribution of initial \(\mathrm{SOC}\) fields and therefore receives \(11{,}000\) samples. As the \acrshort{pefno} additionally learns the effect of varying parameters, the data budget is increased to \(33{,}000\) to populate the four-dimensional parameter space. Each dataset is randomly split into a $90\%/10\%$ training and test set, respectively.

\smallskip
\noindent
Training was done using the Adam optimiser \cite{kingma_adam_2017} with a cosine learning-rate schedule \cite{loshchilov_sgdr_2017}: the rate is linearly warmed from 0 to \(10^{-2}\) during the first epoch, then decays along a cosine curve to \(10^{-4}\) over the remaining epochs, ending with a small step size for fine tuning. All models are implemented in \texttt{JAX} and \texttt{Flax} \cite{bradbury_jax_2018,heek_flax_2024}. The voltage and boundary expressions in Equations~\eqref{eq:spm_c}~and~\eqref{eq:spm_d} are coded in pure JAX, enabling \texttt{jit}-compilation and end-to-end differentiability. Figure \ref{fig:net_c} visualises this step.
\smallskip
\noindent
To avoid overweighting trajectories with larger absolute concentrations, we minimise the \emph{normalised \(\mathrm{L_{2}}\) error}, averaged over the mini-batch:
\begin{equation}
  \mathrm{nL_2}
  \;=\;
  \frac{1}{N}
  \sum_{j=1}^{N}
  \frac{\bigl\lVert\hat{\mathbf c}^{(j)}-\mathbf c^{(j)}\bigr\rVert_{2}}
       {\bigl\lVert\mathbf c^{(j)}\bigr\rVert_{2}},
\end{equation}
where \(N\) is the batch size and \(\hat{\mathbf c}^{(j)},\mathbf c^{(j)}\) are the flattened predicted and reference concentration fields of sample \(j\) respectively. Concentrations are linearly scaled to \([0,1]\) and currents to \([-1,1]\) before training begins.

\subsection{Bayesian identification of electrode diffusivities}

Finally, we embed the \acrshort{pefno} in a Bayesian parameter-identification loop to demonstrate its utility for inverse problems.
We emphasize that the core contribution of this work is the systematic assessment of the surrogate's accuracy, generalisability, and run-time performance; refining its inverse-problem capabilities is a logical next step once the emulator reliably reproduces the full state space of the \acrshort{spm}. Nevertheless, an “out-of-the-box” Bayesian estimation experiment is included here to provide a preliminary analysis of how well the trained operator supports downstream inference tasks and to assess the accuracy of forward vs. inverse inference.

\smallskip
\noindent
Let
\(\boldsymbol\rho=(\log_{10}D_{\mathrm n},\,\log_{10}D_{\mathrm p})\in
\Omega=[-18,-14]^2\subset\mathbb R^{2}\) represent the unknown log‐diffusivities.
For a given current profile \(I(t)\) and initial state
\(c_{0,k}(r)\), the trained \acrshort{pefno} returns the full spatio-temporal concentration field
\(c_k(r,t;\boldsymbol\rho)\).
From this, the electrode surface concentrations are used to compute the cell voltage $V_{\text{pred}}(t;\boldsymbol\rho)$, through Equation \ref{eq:spm_c}.
\noindent
The inverse problem is posed as a bounded minimisation of the relative $\mathrm{L_2}$ error:
\begin{equation}
  \mathrm{nL_2}(\boldsymbol\rho)=
  \frac{\|V_{\text{pred}}(\,\cdot\,;\boldsymbol\rho)-V_{\text{data}}\|_{2}}
       {\|V_{\text{data}}\|_{2}},
\end{equation}
which is evaluated on the discrete voltage trace.
A Gaussian-process surrogate \(m(\boldsymbol\rho)\) is initialised with \(n_{0}=12\) Sobol points in \(\Omega\) and sequentially refined by choosing the next sample
\(\boldsymbol\rho^{\ast}\) as the maximiser of the expected
improvement acquisition function. The optimisation loop terminates after
\(n_{\text{tot}}=60\) evaluations.
The best candidate
\(\boldsymbol\rho_{\!\min}=\arg\min\mathrm{nL_2}(\boldsymbol\theta)\)
provides the point estimate of
\((D_{\mathrm n},D_{\mathrm p})\). The same procedures are repeated for the classical simulation with \texttt{PyBaMM} and compared against each other.

\section{Results}
\label{sec:results}

This section first quantifies the forward accuracy of the three surrogates, then benchmarks their runtime performance, and finally demonstrates how the \acrshort{pefno} supports a Bayesian diffusivity-estimation task.

\subsection{Accuracy of the learned operators}
Added on top of the $\mathrm{nL_2}$, three complementary metrics are reported (explicit formulae are given in the \nameref{app}). Each metric communicates a similar but distinct idea about the prediction accuracy:

\begin{itemize}[leftmargin=1.6em,itemsep=2pt]
\item \textbf{Relative $L_{2}$ error (\(\mathrm{nL_{2}}\)).}  
      Already used for the training, this error scales by the value of the ground truth, giving a clear metric of learning capability irrespective of magnitude.
\item \textbf{RMSE.}
      Root-mean-square error in \si{\milli\volt}; penalises
      larger deviations more strongly than MAE and is widely used in
      battery-model benchmarking.
\item \textbf{MAE.}
      Average absolute error in \si{\milli\volt}; offers an intuitive “per-sample”
      discrepancy that is less influenced by outliers than RMSE.
\item \textbf{Relative $L_{\infty}$ error ($\mathrm{nL_{\infty}}$).}  
      Ratio of the worst-case pointwise error to the maximum reference
      value. This is especially important for detecting peak mismatches that may violate
      safety margins.
\end{itemize}

\begin{figure*}[htb]
  \centering
  \subcaptionbox{}{\includegraphics[width=\textwidth]{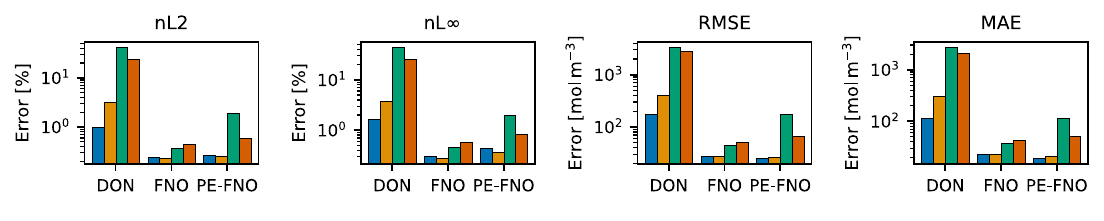}\label{fig:fig3a}}
  \\[0pt]   
  \subcaptionbox{}{\includegraphics[width=\textwidth]{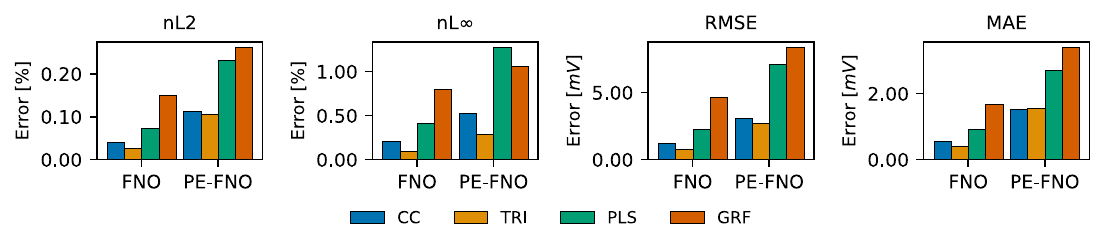}\label{fig:fig3b}}
  \caption{Error metrics for (a) concentration and (b) voltage}
  \label{fig:fig3}
\end{figure*}

Together, these metrics provide a comprehensive view of model performance: unit-aware deviations (RMSE,~MAE), normalised averages (\(\mathrm{nL_{2}}\)), and worst-case behaviour ($\mathrm{nL_{\infty}}$). For concentration errors in Figure~\ref{fig:fig3}, metrics are computed separately for the anode and cathode and then averaged to obtain a single figure.

\medskip \noindent
During data generation, we \emph{did not} enforce the usual \acrshort{spm} safety constraints (e.g., cut-off voltages, admissible stoichiometries, etc.). As a result, a small subset of the simulated trajectories drift marginally outside the real application domain—e.g.\ surface concentrations
exceeding their respective maximum and minimum concentration limits or voltages crossing the \SI{2.5}{\volt}/\SI{3.65}{\volt}
limits. We note that allowing this can have effects on saturation and degradation dynamics due to their path dependencies with the fundamental equation variables. This should be taken into account when one extends this work to models that consider more complex or intricate dynamics than the vanilla \acrshort{spm}.

\medskip
\noindent
Nevertheless, these borderline cases were deliberately retained in the
\emph{training} pool for two reasons:

\begin{enumerate}[leftmargin=1.6em,itemsep=2pt]
\item \textbf{Edge coverage.}  
      Samples that lie just beyond the admissible manifold provide support points near its boundary and teach the surrogate how the solution behaves as the system approaches extreme states.
\item \textbf{Regularisation by extrapolation.}  
      Exposure to slightly out-of-domain inputs encourages the network to learn a smooth continuation of the operator, which improves interpolation in the physical region.
\end{enumerate}

For the accuracy study, these out-of-domain trajectories are removed from the test set, ensuring that all reported metrics correspond strictly to valid operating conditions. We emphasize that the trade-off is that a model trained this way has, on average, a higher error when it is later evaluated \emph{only} on clean data than a model that was exposed to clean data exclusively. I.e., we allow the model to perform worse on average in order to cover all in-domain space at good accuracy. It is for this reason that the $\mathrm{nL_{\infty}}$ error provides a valuable insight into model capabilities.

\paragraph{Concentration fields (Figure \ref{fig:fig3}a)}
\acrshort{deeponet} shows the largest discrepancies, with
$\mathrm{nL_{2}}$ =\,$1.0\,\%$ even for the simple CC profile, with this increasing to $42\,\%$ (RMSE\,$\approx3.3{\times}10^{3}\;\si{\mol\;m^{-3}}$)
for the highly intermittent PLS excitation.
Replacing \acrshort{deeponet} with an \acrshort{fno} reduces the error by
nearly two orders of magnitude. Across all four current families,
$\mathrm{nL_{2}}$ remains below $0.46\,\%$ and the worst-case
$\mathrm{nL_{\infty}}$ does not exceed $0.57\,\%$, while also enabling varying \acrshortpl{soc}.
Adding the parameter-embedding (\acrshort{pefno}) keeps low errors for CC and TRI profiles (e.g.\ $0.27\,\%$ versus
$0.24\,\% \;\mathrm{nL_2}$),
while raiseing PLS and GRF errors to $0.58\,\%$ and $1.9\,\%$, respectively. Even so, \acrshort{pefno} retains a ten-fold advantage over \acrshort{deeponet} while
delivering parameter awareness that the plain \acrshort{fno} lacks.

\paragraph{Voltage traces (Figure \ref{fig:fig3}b)}
Voltage errors follow the same ranking. \acrshort{deeponet} bars are omitted
from the plot for clarity, focusing on the two neural operator variants.
The basic \acrshort{fno} achieves sub-\(0.15\,\%\) $\mathrm{nL_{2}}$ for CC,
TRI and PLS profiles and \(0.15\text{–}0.45\,\%\) $\mathrm{nL_{\infty}}$.
\acrshort{pefno} incurs a factor-of-two penalty across the board
($\mathrm{nL_{2}} \approx 0.11\text{–}0.26\,\%$), resulting in a MAE of
$1.5\text{–}3.4\,\si{mV}$—further illustrating the trade-off for accommodating parameter
variation.

\medskip
\noindent
Overall, \acrshort{fno}-based surrogates achieve sub-percent relative errors for both concentration and voltage. The basic \acrshort{fno} is most accurate when diffusivity is fixed, whereas the parameter-embedded version offers robustness to parameter shifts at the cost of a modest increase in prediction error. \acrshort{deeponet} trails by one to two orders of magnitude across all metrics and current families, underscoring the advantage of neural operator architectures for physics-rich battery problems.

\begin{figure*}[htb]
    \includegraphics[width=\textwidth]{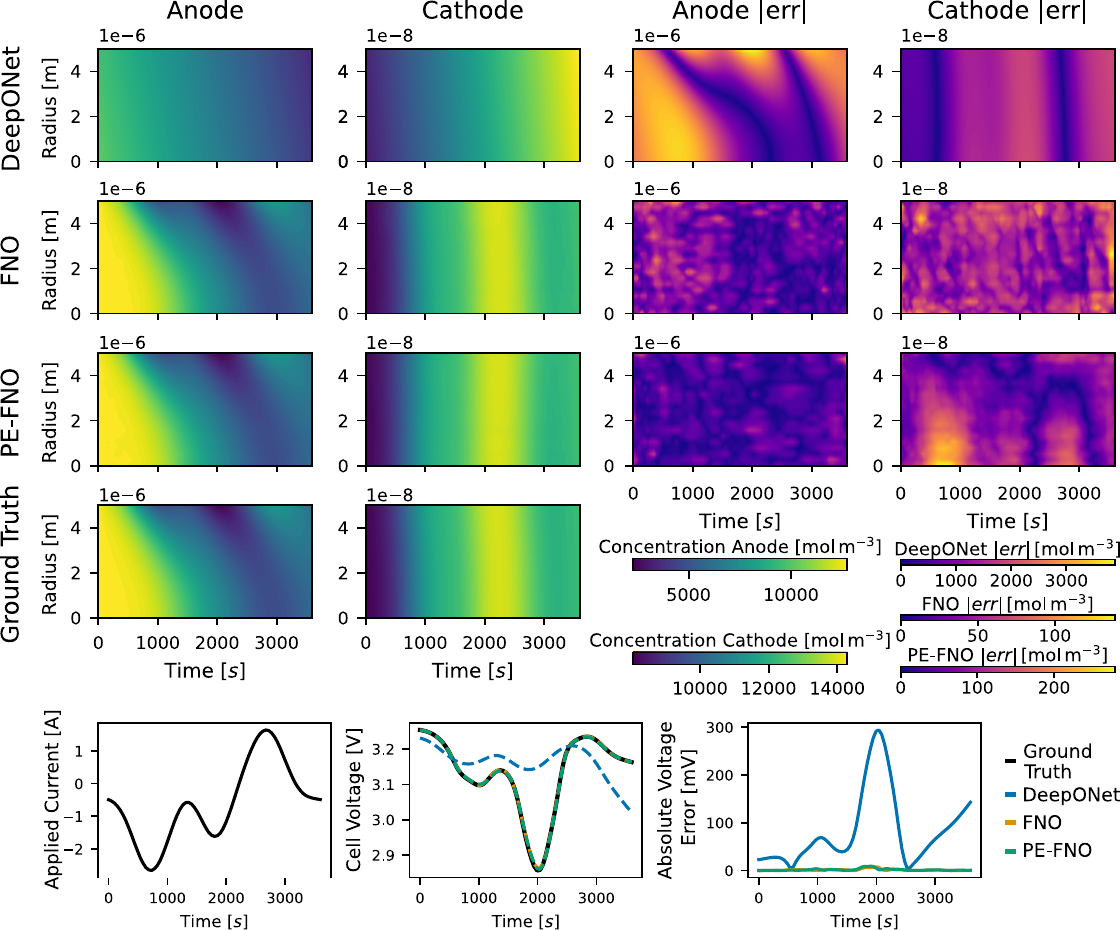}
\caption{Randomly sampled result for a GRF current input.  
Rows: \acrshort{deeponet}, \acrshort{fno}, \acrshort{pefno}, ground-truth \acrshort{spm}.  
Columns: $c_{\mathrm n}(r,t)$, $c_{\mathrm p}(r,t)$ and their absolute
errors; bottom panels give current, voltage, and voltage error.
\acrshort{fno} and \acrshort{pefno} track the \acrshort{spm} within single digit millivolt range,
whereas \acrshort{deeponet} shows spatial artefacts and errors up to
\SI{300}{mV}.}\label{fig:fig4}
\end{figure*}

\smallskip
\noindent
Simulation versus ground truth for one sample of the GRF is visualised in Figure~\ref{fig:fig4}. Here, all three surrogates are queried at the \emph{same} initial
\acrshort{soc} and material parameters; the less capable architectures are therefore trained at the states and parameters they can not attend to.

\subsection{Speed benchmark}\label{sec:speed}

All timings were obtained on a workstation equipped with an AMD Ryzen Threadripper PRO 5965WX CPU (48 physical cores) and an NVIDIA RTX A4000 GPU. The configuration reflects a practical deployment scenario where the \acrshort{spm} is executed on the CPU while the learned operators run on the GPU.

\paragraph{CPU baseline}
The \acrshort{spm} was solved using \texttt{PyBaMM} v\texttt{24.9.0} with the solver from \citet{andersson_casadi_2019}. Runs were repeated for \(n_{\text{thr}}\in\{1,8,16\}\) threads, and reported
figures correspond to wall–clock time divided by the number of trajectories in the batch, such that every entry represents the average runtime of a \emph{single} simulation.

\paragraph{GPU surrogates}
All neural operators were benchmarked with a fixed inference batch of \(B=100\) samples, the largest size that fits the \acrshort{pefno} into the RTX-A4000’s \SI{16}{GB} VRAM. The raw batch time was divided by \(B\) to obtain a per-trajectory latency directly comparable to the CPU baseline.

\begin{figure*}[htb]
    \centering
    \includegraphics[width=\columnwidth]{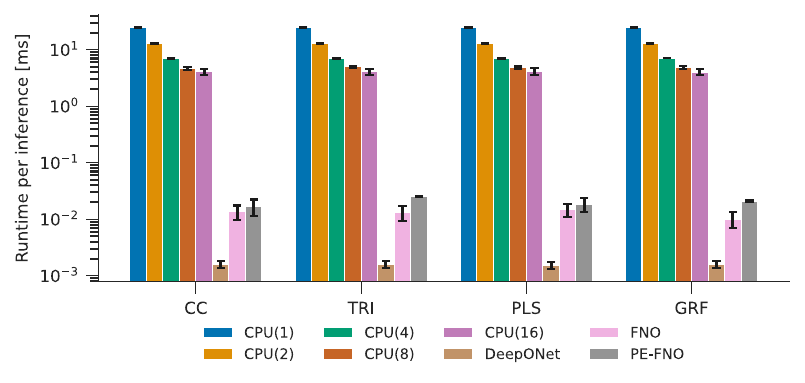}
\caption{Speed comparison between learned operators and classical solver parallelised to different numbers of cores; all values normalised to a per simulation run basis}\label{fig:fig5}
\end{figure*}

Figure~\ref{fig:fig5} shows that

\begin{itemize}[leftmargin=1.6em,itemsep=2pt]
\item \textbf{\acrshort{deeponet}} achieves the lowest latency,
      \(\approx 1.6~\si{\micro\second}\) per trajectory,
      delivering a \(>\!3{,}500\times\) speed-up over the CPU baseline.
\item \textbf{\acrshort{fno}} completes one forward pass in
      \(10\text{–}15~\si{\micro\second}\); the FFT and complex
      multiplications introduce a $6-9$ times overhead relative to DeepONet
      but still yield a \(>400\times\) speed-up over the numerical
      baseline.
\item \textbf{\acrshort{pefno}} adds a further \(20\!-\!40\,\%\) cost for the
      parameter pre-processing, but remains two orders of magnitude faster than the numerical SPM and is well suited for real-time or embedded
      applications.
\end{itemize}

\noindent
In practice, all surrogates run in sub-milliseconds, enabling latency-critical tasks such as embedded state estimation and inner-loop optimal control.

\medskip 
\noindent
A fair speed comparison must include the \emph{up-front cost} of
producing each surrogate. Table \ref{tab:train-cost} reports the data-generation time and training time needed
to reach the test accuracies in Figure~\ref{fig:fig3}.
\smallskip
\noindent
Training a \acrshort{deeponet} requires minimal time ($\sim\SI{1}{min}$), whereas the full
\acrshort{pefno} takes approximately \SI{1.5}{h} on a single RTX-A4000. Contextualising this with the simulation benchmarks from Figure \ref{fig:fig5} leads to the conclusion that  \acrshort{pefno} becomes advantageous only when the \acrshort{spm} must
be evaluated on the order of a million times. A use setting that occurs in typical battery management systems or model predictive control applications.

\begin{table}[htbp]
  \centering
  \caption{Data volume and wall-clock cost for training each surrogate on a
           single RTX-A4000.  All times are approximate.}
  \label{tab:train-cost}
  \setlength{\tabcolsep}{4pt}       
  \begin{tabular}{l
                  S[table-format=2]
                  l
                  l}
    \toprule
    Model &
    {\#\,epochs} &
    Training time &
    Generation time\\
    \midrule
    \acrshort{deeponet} & 15 & $\approx\!\SI{1}{min}$   & $\approx\!\SI{2}{min}$  \\
    \acrshort{fno}      & 25 & $\approx\!\SI{20}{min}$  & $\approx\!\SI{11}{min}$ \\
    \acrshort{pefno}   & 50 & $\approx\!\SI{60}{min}$     & $\approx\!\SI{33}{min}$ \\
    \bottomrule
  \end{tabular}
\end{table}

\subsection{Inversion capabilities of the \acrshort{pefno}}

\begin{figure}[htb]
  \makebox[\columnwidth]{%
    \includegraphics{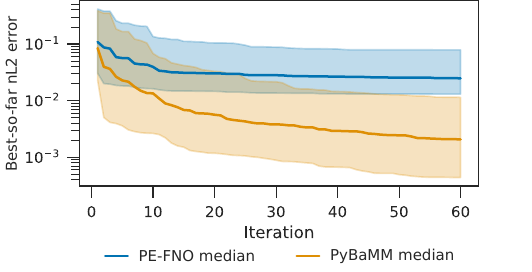}
  }
\caption{Convergence plot for the voltage fit for estimating $D^*_n$ and $D^*_p$ from data. Median in thick line and $25-75\%$ ranges over $1000$ samples}\label{fig:fig6}
\end{figure}

Figure \ref{fig:fig6} plots the best-so-far voltage loss (logarithmic) during Bayesian
optimisation of \(\log_{10}D_{\mathrm n}\) and
\(\log_{10}D_{\mathrm p}\) for all $1000$ samples out of the test set.  

\texttt{PyBaMM} converges to a plateau of
\(\mathrm{nL_{2}}\approx10^{-3}\) after approximately $35$ evaluations, whereas
PE-FNO settles at \(\mathrm{nL_{2}}\approx2{\times}10^{-2}\) within
15 iterations.  The gap equals a voltage deviation of
\(0.049\,\si{mV}\) versus \(0.56\,\si{mV}\) in the RMSE for the depicted median, i.e., an approximately one-decade loss in inverse accuracy for the surrogate.
\smallskip
\noindent
Across the complete sample set, the mean voltage error is
\(\mathrm{nL_2}=0.157\,\%\) for \texttt{PyBaMM} and
\(0.491\,\%\) for the \acrshort{pefno}, i.e.\ a difference of
\(0.334\) percentage points.  
The surrogate’s inverse error is thus about
\(27\,\%\) higher than its forward‐prediction error
(\(\mathrm{nL_2}=0.262\,\%\)), consistent with the fact that the neural operator
was only trained on the forward map.

\begin{table}[htbp]
  \centering
  \caption{Diffusivity-estimate error metrics.}
  \label{tab:diffusivity-metrics}

  \begin{tabular}{%
      ll
      S[table-format=1.4]
      S[table-format=1.4]
      S[table-format=1.4]
    }
    \toprule
    {Parameter} & {Model} & {MAE} & {RMSE} & {MAPE [\%]} \\
    \midrule
    \multirow{2}{*}{$\log_{10}D_{an}$}
      & PE-FNO  & 0.1465 & 0.2524 & 1.1432 \\
      & PyBaMM  & 0.0191 & 0.0286 & 0.1428 \\
    \midrule
    \multirow{2}{*}{$\log_{10}D_{ca}$}
      & PE-FNO  & 1.1408 & 1.3971 & 8.4783 \\
      & PyBaMM  & 1.1163 & 1.3685 & 8.2947 \\
    \bottomrule
  \end{tabular}
\end{table}

\paragraph{Recovered parameters}
Table \ref{tab:diffusivity-metrics} lists the resulting diffusivity
errors.  
For the anode, \acrshort{pefno} attains an \(\mathrm{RMSE}=0.25\) dex; eight times less
precise than \texttt{PyBaMM} but achieved at millisecond latency.  
For the cathode, both estimators are limited to
\(\approx1.4\) dex due to the flat LiFePO\(_4\) open-circuit potential,
confirming that voltage carries little information about
\(D_{\mathrm p}\) for this chemistry. 

\section{Conclusion and Discussion}
\label{sec:conc}

In this work, multiple operator-learning architectures were benchmarked for their capabilities of learning the \acrshort{spm}'s solution operator across varying input currents, initial \acrshortpl{soc} and parameter values.
\acrshort{deeponet} was able to learn the solutions for CC and TRI current inputs at a fixed \acrshort{soc} and parameter set, while the \acrshort{fno} extended this capability to PLS and GRF currents and varying \acrshortpl{soc}, marking the first application of a discretisation-invariant neural operator in lithium-ion battery modelling. Furthermore, a lightweight parameter-embedding module supplies channel-wise
scale–shift factors derived from particle radius and solid-phase diffusivities, enabling a single network \acrshort{pefno} to cope simultaneously with arbitrary
\acrshort{soc} fields, highly dynamic current profiles, and a two-decade range of material parameters.

\smallskip
\noindent
Compared with an average multithreaded \texttt{PyBaMM} \acrshort{spm} solver, the trained \acrshort{pefno} is approximately two orders of magnitude faster while still delivering sub-percent errors in both concentration and voltage predictions. Importantly, because the model is mesh-agnostic, the \emph{same weights} can be deployed on coarse grids for
rapid simulation or on fine grids for high-resolution analysis without
retraining. This is an area that warrants further exploration in future work.

\smallskip
\noindent
Embedding the operator in a Bayesian optimisation loop further revealed the expected trade-off between forward fidelity and inverse accuracy: the surrogate recovers estimated diffusivity values at one decade higher error than \texttt{PyBaMM}, but with approximately two decades faster runtime.

\smallskip
\noindent
Neural-operator surrogates are most valuable in workflows that require \emph{extensive} simulator sampling—state estimation, real-time control, and design studies—where the cost of generating the training set quickly dwarfs the number of simulation runs needed during deployment.
For such applications, it is essential to view a surrogate’s merit holistically: forward accuracy, inverse accuracy, and execution speed must be weighed against training effort, data-generation cost, and, above all, generalisability. By reporting each of these factors side-by-side, the present study paves the way for fair assessments of physics-inspired machine learning frameworks. We conclude with the remark that the purely data-driven \acrshort{pefno} offer a much more favourable balance between training time and accuracy than \acrshortpl{pinn} and also generalises more robustly across parameters, initial states, and boundary conditions.

Future work should focus on expanding the physical envelope of these surrogates, including extending them to additional parameters and more complex electrochemical models (e.g., \acrshort{dfn}). This would capture a wider set of conditions relevant to battery management systems. Likewise, a single operator that copes with constant-current, pulse and stochastic load
families would remove the need for family-specific training. Finally,
incorporating inverse awareness through joint forward–adjoint training offers a promising route to narrowing the remaining
gap in parameter-estimation accuracy while retaining the speed advantage demonstrated here.

\section*{CRediT authorship contribution statement}

\textbf{Amir Ali Panahi}: Conceptualization, Methodology, Software, Validation, Investigation, Writing – original draft, Writing – review \& editing, Visualization.\\
\textbf{Daniel Luder}: Conceptualization, Methodology, Writing – review \& editing, Supervision.\\
\textbf{Billy Wu}: Supervision, Writing – review \& editing, Funding acquisition.\\
\textbf{Gregory Offer}: Writing – review \& editing.\\
\textbf{Dirk Uwe Sauer}: Writing – review \& editing.\\
\textbf{Weihan Li}: Supervision, Writing – review \& editing, Funding acquisition.

\section*{Code availability}
The code will be made publicly available upon publication of the manuscript.

\section*{Declaration of Competing Interest}

The authors declare that they have no known competing financial
interests or personal relationships that could have appeared to influence the work reported in this paper.

\section*{Acknowledgements}
This work has received funding from the project "SPEED" (03XP0585) funded by the German Federal Ministry of Research, Technology and Space (BMFTR) and the project "ADMirABLE" (03ETE053E) funded by the German Federal Ministry for Economic Affairs and Energy (BMWE). The authors thank the support of Shell Research UK Ltd. for the Ph.D. studentship of Amir Ali Panahi and the EPSRC Faraday Institution Multi-Scale Modelling Project (FIRG084).

\printglossary[type=\acronymtype]

\printglossary

\appendix
\section*{Appendix}
\phantomsection
\label{app}
\makeatletter
\renewcommand{\thetable}{\arabic{table}}
\renewcommand{\p@table}{}
\makeatother
\subsection*{Parameter Values}
\label{app:params}

\begin{table*}[!ht]
\centering
\caption{Model parameters for the cell of \cite{chen_development_2020} and the cell of \cite{prada_simplified_2013}.}
\resizebox{\textwidth}{!}{%
\begin{tabular}{l l c c l}
\hline
\textbf{Symbol} & \textbf{Name} &
\textbf{Chen2020 (NMC)} & \textbf{Prada2013 (LFP)} &
\textbf{Unit}\\
\hline
$C$                         & Capacity                                            & 5           & 2.3          & Ah\\
$c_{n,\max}$                & Max.\ conc.\,neg.\ el.                                 & 33133       & 30555        & mol\,m$^{-3}$\\
$c_{p,\max}$                & Max.\ conc.\,pos.\ el.                                 & 63104       & 22806        & mol\,m$^{-3}$\\
$U_{eq,n}$                  & OCP\,neg.\ electrode                                   & \ref{eq:ocpn_chen} & \ref{eq:ocpn_chen} & V\\
$\alpha_n$                  & Charge-transfer coeff.\,neg.\ el.                      & 0.5         & 0.5          & –\\
$\sigma_n$                  & Conductivity\,neg.\ el.                                & 215         & 215          & S\,m$^{-1}$\\
$\varepsilon_n$             & Porosity\,neg.\ el.                                    & 0.25        & 0.36         & –\\
$L_n$                       & Thickness\,neg.\ el.                                   & 8.52$\times10^{-5}$ & 3.4$\times10^{-5}$ & m\\
$D_n$                       & Diffusivity\,neg.\ particle                            & 3.3$\times10^{-14}$ & 3$\times10^{-15}$ & m$^{2}$\,s$^{-1}$\\
$R_n$                       & Radius\,neg.\ particle                                 & 5.86$\times10^{-6}$ & 5$\times10^{-6}$ & m\\
$U_{eq,p}$                  & OCP\,pos.\ electrode                                   & \ref{eq:ocpp_chen} & \ref{eq:ocpp_afshar2017} & V\\
$\alpha_p$                  & Charge-transfer coeff.\,pos.\ el.                      & 0.5         & 0.5          & –\\
$\sigma_p$                  & Conductivity\,pos.\ el.                                & 0.18        & 0.338        & S\,m$^{-1}$\\
$\varepsilon_p$             & Porosity\,pos.\ el.                                    & 0.335       & 0.426        & –\\
$L_p$                       & Thickness\,pos.\ el.                                   & 7.56$\times10^{-5}$ & 8$\times10^{-5}$ & m\\
$D_p$                       & Diffusivity\,pos.\ particle                            & 4$\times10^{-15}$ & 5.9$\times10^{-18}$ & m$^{2}$\,s$^{-1}$\\
$R_p$                       & Radius\,pos.\ particle                                 & 5.22$\times10^{-6}$ & 5$\times10^{-8}$ & m\\

\hline
\end{tabular}}
\label{tab:cell_parameters}
\end{table*}

Let $\mathrm{sto}$ be the normalised concentration at the respective particle, then $U^p_{OCP}$ respective $U^n_{OCP}$ for \cite{chen_development_2020} are defined by:
\begin{align}\label{eq:ocpp_chen}
    U^p_{OCP}(\mathrm{sto}) = &-0.8090 \, \mathrm{sto}
+ 4.4875 \\
&- 0.0428 \tanh\bigl(18.5138(\mathrm{sto} - 0.5542)\bigr)\nonumber \\
&- 17.7326 \tanh\bigl(15.7890(\mathrm{sto} - 0.3117)\bigr)\nonumber \\
&+ 17.5842 \tanh\bigl(15.9308(\mathrm{sto} - 0.3120)\bigr)\nonumber .
\end{align}

\begin{align}\label{eq:ocpn_chen}
    U^n_{OCP}(\mathrm{sto}) = &+1.9793 \exp(-39.3631 \,\mathrm{sto})
+ 0.2482\\
&- 0.0909 \tanh\bigl(29.8538(\mathrm{sto} - 0.1234)\bigr)\nonumber \\
&- 0.04478 \tanh\bigl(14.9159(\mathrm{sto} - 0.2769)\bigr)\nonumber \\
&- 0.0205 \tanh\bigl(30.4444(\mathrm{sto} - 0.6103)\bigr)\nonumber .
\end{align}

The positive electrode OCP from \cite{prada_simplified_2013} is given by:
\begin{align}\label{eq:ocpp_afshar2017}
    U^p_{OCP}(\mathrm{sto}) = &+3.4077 - 0.020269 \,\mathrm{sto} \\
    &+ 0.5 \exp(-150\,\mathrm{sto})
    - 0.9 \exp(-30(1 - \mathrm{sto}))\nonumber .
\end{align}

\subsection*{Current families}
\label{app:currents}
Let $T_{\max}$ be the experiment horizon and $n$ the temporal
resolution used throughout the data set.
Define the uniform grid
\[
   \mathcal{T}
     =\bigl\{t_i\bigr\}_{i=1}^{n},
   \qquad
   t_i \;=\; \frac{i-1}{n-1}\,T_{\max}.
\]

\paragraph{(a) Constant-current family.}
\[
   I(t) = I_{\mathrm{const}},\qquad t\in[0,T_{\max}],
\]
with $|I_{\mathrm{const}}|\le 1.5C$.

\paragraph{(b) Triangular current family.}
Choose $0<t_1<t_2\le T_{\max}$ and a peak value
$I_{\triangle}\in[-1.5C,1.5C]$; then
\[
   I(t)=
   \begin{cases}
     \displaystyle I_{\triangle}\,\dfrac{t}{t_1}, & 0\le t\le t_1,\\[8pt]
     \displaystyle I_{\triangle}\,\dfrac{t_2-t}{t_2-t_1}, & t_1<t\le t_2,\\[8pt]
     0, & t>t_2.
   \end{cases}
\]

Setting $t_1=1800\;\si{s}$ and $t_2=3600\;\si{s}$ reproduces the profile
used in the numerical experiments.

\paragraph{(c) Rectangular-pulse family.}
A random pulse train $I_{\mathrm{rect}}(t)$ is generated as follows.
Let $T_{\max}$ be the horizon of the experiment and $C$ the 1\,C rate.

\begin{enumerate}
\item \emph{Pulse count.}  
      Draw an integer
      \(
         N_h \sim \mathrm{Unif}\{1,\dots,10\}
      \)
      (pulses per hour) and set
      \[
         n_p \;=\;
         \max\!\Bigl\{1,\,
           \bigl\lfloor N_h\,T_{\max}/3600\bigr\rfloor
         \Bigr\}.
      \] for the total number of pulses
\item \emph{Amplitude.}  
      Select a direction $s\in\{+1,-1\}$ at random and a magnitude factor
      $a\sim\mathcal{U}(0.2,1.5)$; the amplitude equates to:
      \[
         I_\mathrm{p}\;=\;s\,a\,C,\qquad |I_\mathrm{p}|\in[0.2C,1.5C].
      \]
\item \emph{Timing.}  
      The period is
      $P=T_{\max}/n_p$.
      Draw a duty cycle
      $d\sim\mathcal{U}(0.2,0.7)$ and set the pulse width
      $\tau=dP$.
      Pulse $k$ starts at $t_k=kP$ for $k=0,\dots,n_p-1$.
\end{enumerate}

\noindent
The resulting current is
\[
   I(t)=
   \begin{cases}
     I_\mathrm{p}, &\displaystyle
       \exists\,k\!:\; 0\le k<n_p,\quad
       t\in[t_k,\;t_k+\tau),\\[6pt]
     0, & \text{otherwise}.
   \end{cases}
\]

\paragraph{(d) Gaussian–random-field (GRF) family.}
For a length scale $L=1$ we adopt the periodic squared-exponential
covariance
\begin{equation}
   k_{\text{per}}(t,t')
   \;=\;
   \exp\!\bigl[
      -2\,\sin^2\!\bigl(\tfrac{\pi}{T_{\max}}(t-t')\bigr)\bigm/ L^{2}
   \bigr],
   \quad t,t'\in[0,T_{\max}],
   \label{eq:per_se}
\end{equation}
which guarantees $I(0)=I(T_{\max})$ in distribution and represents loosely a cycling protocol.  
Let
\[
   \boldsymbol{K}\;=\;
      \bigl[k_{\text{per}}(t_i,t_j)\bigr]_{i,j=1}^{n}
      \;+\;
      \varepsilon^{2}\mathbf{I}_n,
   \qquad
      \varepsilon=10^{-3},
\]
and draw
\begin{equation}
   \mathbf{y}\sim\mathcal{N}\!\bigl(\mathbf{0},\,\boldsymbol{K}\bigr).
   \label{eq:grf_sample}
\end{equation}
The continuous current is obtained by linear interpolation between the sampled currents. The current is clipped and scaled to be in the preferred operating range:
\[
   I(t)
   = \operatorname{clip}\bigl(I_{\text{GRF}}(t),-1.5,1.5\bigr)\times C.
\]

\subsection*{Error metrics}
\label{app:errors}

Given prediction \(\hat y\) and ground truth \(y\) on an
\(n_r\times n_t\) grid we report four scalar metrics:

\[
\begin{aligned}
  \mathrm{MAE}      &= \frac{1}{n_r n_t}\sum_{p,q}\lvert\hat y_{pq}-y_{pq}\rvert,
\\[6pt]
  \mathrm{RMSE}      &= \sqrt{\frac{1}{n_r n_t}\sum_{p,q}(\hat y_{pq}-y_{pq})^{2}},
\\[6pt]
  \mathrm{nL_2}&= \frac{\lVert\hat y-y\rVert_{2}}
                             {\lVert y\rVert_{2}+\varepsilon},
  \qquad
  \lVert y\rVert_{2}=\Bigl(\sum_{p,q}y_{pq}^{2}\Bigr)^{1/2},
\\[6pt]
  \mathrm{nL_{\infty}}
                    &= \frac{\lVert\hat y-y\rVert_{\infty}}
                             {\lVert y\rVert_{\infty}+\varepsilon},
  \qquad
  \lVert y\rVert_{\infty}=\max_{p,q}\lvert y_{pq}\rvert,
\end{aligned}
\tag{F2--F5}
\]
with a small constant \(\varepsilon=10^{-12}\) to avoid division by
zero.


 \bibliographystyle{elsarticle-num-names} 
 \bibliography{references.bib}






\end{document}